\pdfoutput=1

\documentclass[11pt]{article}

\usepackage[]{EMNLP2022}

\usepackage{graphicx}
\usepackage{times}
\usepackage{latexsym}
\usepackage{enumitem}
\usepackage{amsmath}
\usepackage{float}

\usepackage[T1]{fontenc}

\usepackage[utf8]{inputenc}

\usepackage{microtype}

\usepackage{inconsolata}

\title{OPD@NL4Opt: An ensemble approach for the NER task of the optimization problem}

\author{Kangxu Wang,\, Ze Chen,\,  Jiewen Zheng\\
         Interactive Entertainment Group of Netease Inc., Guangzhou, China \\
         \texttt{\{wangkangxu,jackchen,zhengjiewen\}@corp.netease.com}}

\begin{document}
\maketitle
\begin{abstract}
In this paper, we present an ensemble approach for the NL4Opt competition subtask 1(NER task). For this task, we first fine tune the pre-trained language models based on the competition dataset. Then we adopt differential learning rates and adversarial  training strategies to enhance the model generalization and robustness. Additionally, we use a model ensemble method for the final prediction, which achieves a micro-averaged  F1 score of 93.3\% and attains the second prize in the NER task.
\end{abstract}

\section{Introduction}
Named Entity Recognition (NER) aims to detecting the boundaries of named entities and recognizing their categories(e.g., person or location). It plays an important role in many downstream tasks, such as information extraction and question answering. In optimization problems, many semantic entities (such as decision variables, objective, constraints) are very helpful for optimization solvers. NLP methods (such as NER) can help automatically translate an optimization problem description into a format that optimization solvers can understand\citep{nl4opt}.

\section{Background}
\subsection{Task Description}
\begin{figure}[htbp]
   \centering
   \includegraphics[width=\linewidth,scale=1.00]{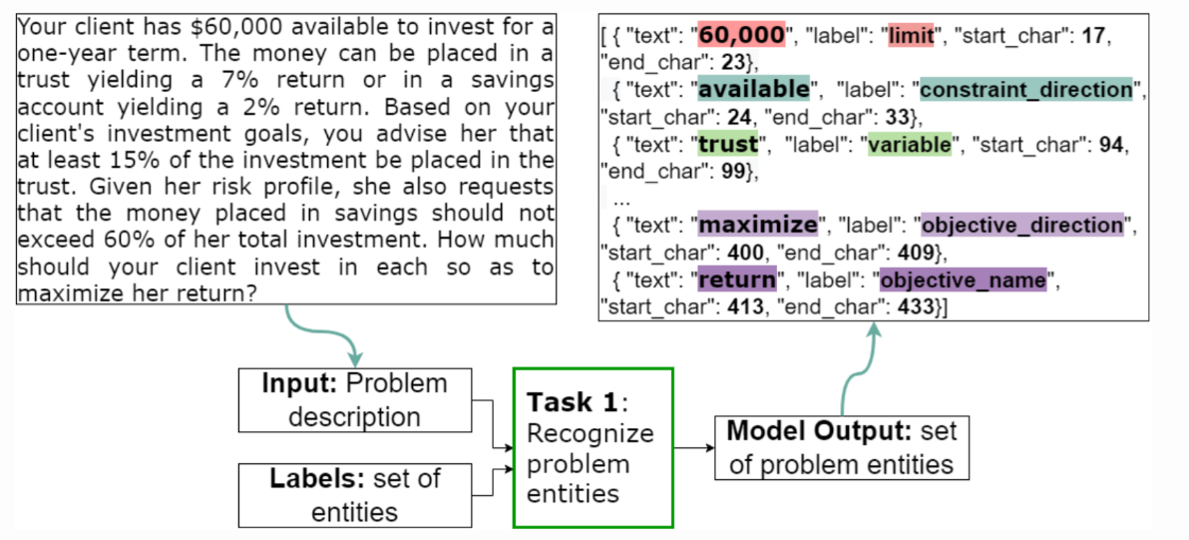} 
   \caption{NL4Opt competition subtask 1}
   \label{task}
\end{figure}

The  goal  of  this task\citep{nl4opt} is to recognize the label the semantic entities that correspond to the components of the optimization problem. It aims to reduce the ambiguity by detecting and tagging the entities of the optimization problems such as the objective name, decision variable names, or the constraint limits. The dataset of NL4Opt contains approximately 1100 annotated linear programming (LP) word problems from 6 different domains. Figure \ref{task} gives us details about this task.

\subsection{Pre-trained Language Models}
Recently, pre-trained language models (PLMs) have achieved remarkable achievement on natural language processing tasks, becoming one of the most effective methods for engineers and scholars. Transformers-based Pre-trained language models such as BERT\cite{devlin2018bert},  RoBERTa\cite{liu2019roberta}, DeBERTa\citep{he2020deberta}, DeBERTaV3\cite{he2021debertav3} is designed to pre-train deep representation from unlabeled text, which can be fine-tuned with just one additional output layer to create state-of-the-art models for a wide range of tasks, such as question answering and language inference, without substantial task-specific architecture modifications.

\section{System Overview}
In this section, we first present the framework details for the models adopted in our work. Then we introduce several strategies for improving the models' robustness. Finally, we talk about the design of the model ensemble method.
\subsection{Model Architecture}
\begin{figure}[htbp]
   \centering
   \includegraphics[width=\linewidth,scale=1.20]{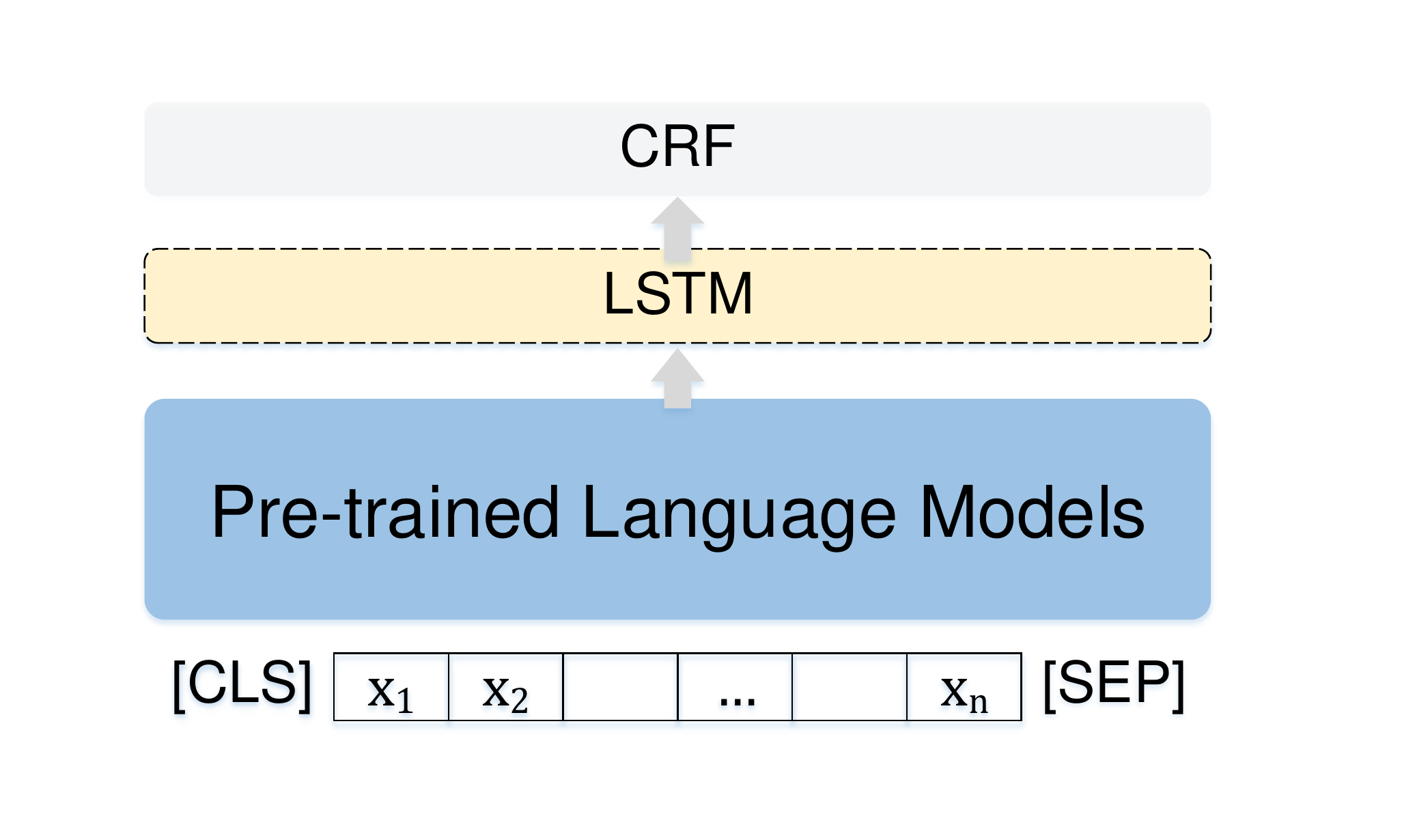} 
   \caption{Model framework}
   \label{moe_model}
\end{figure}

In our experiments, we compare the performance of many BERT-based NER model architecture (see details in Figure \ref{moe_model}), including BERT, BERT+CRF and BERT+LSTM+CRF. In this task, BERT+CRF has a better result than others. We also investigate the impact of adopting different pre-trained LMs, finding that DeBERTa performs best.

\subsection{Adversarial Training}
\begin{figure}[htbp]
   \centering
   \includegraphics[width=\linewidth,scale=1.00]{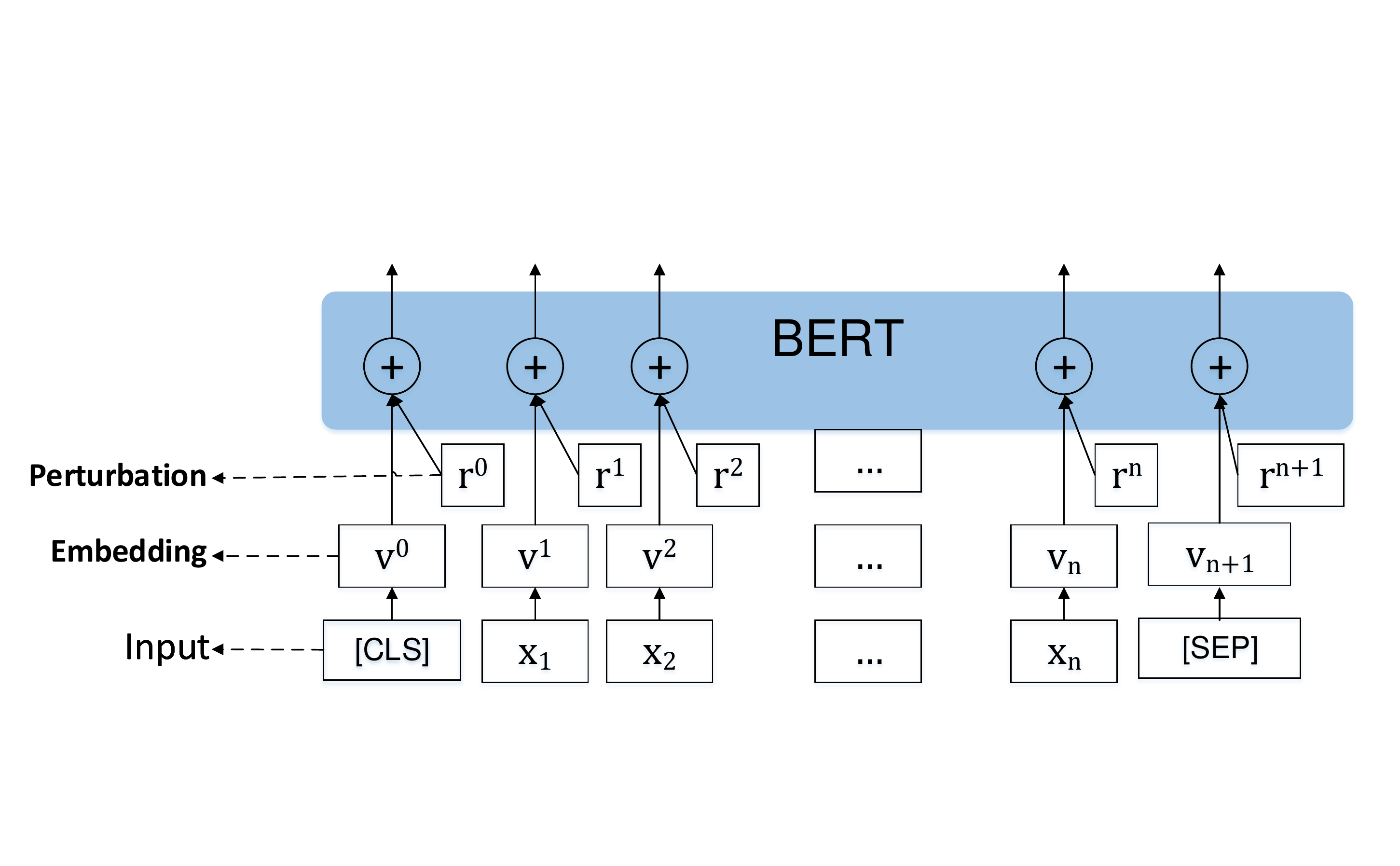} 
   \caption{Adversarial training}
   \label{adv}
\end{figure}

Adversarial attack has been well applied in both computer vision and natural language processing to improve the model's robustness. We implement this strategy with Fast Gradient Method\citep{goodfellow2014explaining}, which directly uses the gradient to compute the perturbation and augments the input with this perturbation to maximizes the adversarial loss. The training procedure can be summarized as follows:
$$\min_{\theta}E_{(x,y)\sim \mathcal{D}}[\max_{\Delta x\in \Omega} L(x+ \Delta x, y; \theta)]$$
where x is input, y is the gold label, $\mathcal{D}$ is the dataset, $\theta$ is the model parameters, $L(x+ \Delta x, y; \theta)$ is the loss function and $\Delta x$ is the perturbation. Figure \ref{adv} gives us a glimpse ofthe  adversarial learning procedure. 

\subsection{Differential learning rates}
After  pre training, PLMs only need a very small learning rate (i.e. 2e-5)  when finetune downstream tasks. If the leaning rate is  too large, it may not  converge well. But  CRF layer need larger learning rate because it is not  pre trained. In our experiment, we Increase the learning rate of CRF layer to 100 times that of PLMs.

\subsection{Model Ensemble}
In this task, the number of  samples in  train dataset is less  than one thousand, therefore, the performance of  model  fluctuates greatly when using different  random  seeds. We train model  with different random seeds. Given predictions  from different  random  seeds, we use majority voting to generate the final prediction. We convert  the label sequences into entity spans to perform majority voting.

\section{Experiments}
\subsection{Experimental Setup}
Our implementation is based on the Transformers library by HuggingFace\citep{wolf2019huggingface} for the pre-trained models and corresponding tokenizers. During training, the data is processed by batches of size 8, the maximum length of each sample is set to 256, and the learning rate is set to 1e-6 with a warmup ratio over 10\%. By default, we set $\epsilon$ to 1.0 in FGM.

\subsection{Results and Analysis}
In this section, we first present experimental results on the base model. Then we experiment with MoE models using the effective strategies validated on the base model. At last, the results of the model ensemble are reported.

We explore the impact of different pre-trained LMs adopted as the contextual encoder. Results given in Table \ref{MoEResults} show that DeBERTa-large can perform well on this task, and monolingual models perform better than multilingual models on this task. By adopting FGM and differential learning rates, the performance can improve a lot.

When we experiment with single models, we find the uneven performance on different semantic entities. Table \ref{WordResults} shows the details, model performs worst on OBJ\_NAME. To make our model generalize well on other dataset, we choose models which perform well on OBJ\_NAME for the final model ensemble process.

Table \ref{EnsembleResults} gives the results of our model ensemble method. Our final prediction achieves a micro-averaged  F1 score of 93.3\% and attains the second prize in the NER task.

\begin{table}\footnotesize
\centering
\begin{tabular}{lcc} 
\hline
\textbf{Model}  & \textbf{micro-F1} & \textbf{macro-F1} \\ 
\hline
xlm-roberta-base + lstm + crf  & 89.18\%            & 89.21\% \\
xlm-roberta-large+lstm+crf  & 90.96\%            & 88.20\% \\
infoxlm-large+lstm+crf  & 91.89\%            & 89.26\% \\
infoxlm-large+lstm+focal-loss   & 91.85\%            & 89.43\% \\
infoxlm-large+crf   & 91.54\%            & 88.76\% \\
deberta-v3-large+lstm+crf  & 91.81\%            & 89.36\% \\
deberta-v3-large+crf  & 92.18\%            & 89.35\% \\
deberta-v3-large+crf+fgm  & 92.45\%            & 89.61\% \\
+5 models ensemble         & 92.61\%            & 90.48\% \\
+9 models ensemble         & 93.05\%            & 90.73\% \\
\hline
\end{tabular}
\caption{Results of different models on dev dataset}
\label{MoEResults}
\end{table}

\begin{table}\footnotesize
\centering
\begin{tabular}{lccc} 
\hline
\textbf{Type}  & \textbf{P} & \textbf{R} & \textbf{F1} \\ 
\hline
CONST\_DIR  & 95.71\%            & 91.39\%  & 93.50\%\\
LIMIT  & 89.13\%            & 89.13\%  & 89.13\%\\
OBJ\_DIR  & 100.00\%            & 100.00\%  & 100.00\%\\
OBJ\_NAME  & 79.73\%            & 59.90\%  & 58.64\%\\
PARAM  & 96.76\%            & 99.33\%  & 98.03\%\\
VAR  & 94.57\%            & 96.12\%  & 95.34\%\\
\hline
\end{tabular}
\caption{Results of different semantic types}
\label{WordResults}
\end{table}

\begin{table}[H]
\centering
\begin{tabular}{ccc} 
\hline
\textbf{model}  & \textbf{F1 score}  \\ 
\hline
deberta-v3-large+crf+fgm(ensemble)    & \bf{93.3\%}      \\
\hline
\end{tabular}
\caption{Ensemble results on Test dataset}
\label{EnsembleResults}
\end{table}

\section{Conclusion}
In this work, we provide an overview of the combined approach to recognize entities in NL4Opt .We investigate the impact of adopting
different pre-trained LMs, finding that DeBERTa performs best in  this task. Experimental results show  that  strategies  such as larger learning rate for CRF layer, adversarial training and model ensembling  can enhance  the model’s effectiveness.


\bibliography{anthology,custom}
\bibliographystyle{acl_natbib}

\end{document}